\documentclass[letterpaper]{article} 
\usepackage{aaai2026}  
\usepackage{times}  
\usepackage{helvet}  
\usepackage{courier}  
\usepackage[hyphens]{url}  
\usepackage{graphicx} 
\usepackage{natbib}  
\usepackage{caption} 
\usepackage{algorithm}
\usepackage{algorithmic}

\usepackage{amsmath}
\usepackage{amsfonts}
\usepackage{array}
\usepackage{tabularx}
\usepackage{subcaption}
\usepackage[dvipsnames]{xcolor}

\usepackage{newfloat}
\usepackage{listings}
\DeclareCaptionStyle{ruled}{labelfont=normalfont,labelsep=colon,strut=off} 
\floatstyle{ruled}
\newfloat{listing}{tb}{lst}{}
\floatname{listing}{Listing}
\title{KTCF: Actionable Recourse in Knowledge Tracing via Counterfactual Explanations for Education}

\author{
    Woojin Kim,
    Changkwon Lee,
    Hyeoncheol Kim
}
\affiliations{
    Department of Computer Science and Engineering, Korea University\\
    \{woojinkim1021, eckdrnjs, harrykim\}@korea.ac.kr
}

\begin{document}

    \maketitle

\begin{abstract}
Using Artificial Intelligence to improve teaching and learning benefits greater adaptivity and scalability in education. 
Knowledge Tracing (KT) is recognized for student modeling task due to its superior performance and application potential in education. 
To this end, we conceptualize and investigate counterfactual explanation as the connection from XAI for KT to education. 
Counterfactual explanations offer actionable recourse, are inherently causal and local, and easy for educational stakeholders to understand who are often non-experts. 
We propose {\tt KTCF}, a counterfactual explanation generation method for KT that accounts for knowledge concept relationships, and a post-processing scheme that converts a counterfactual explanation into a sequence of educational instructions. 
We experiment on a large-scale educational dataset and show our {\tt KTCF} method achieves superior and robust performance over existing methods, with improvements ranging from 5.7\% to 34\% across metrics.
Additionally, we provide a qualitative evaluation of our post-processing scheme, demonstrating that the resulting educational instructions help in reducing large study burden.
We show that counterfactuals have the potential to advance the responsible and practical use of AI in education.
Future works on XAI for KT may benefit from educationally grounded conceptualization and developing stakeholder-centered methods. 
\end{abstract}

\section{Introduction}

The use of Artificial Intelligence (AI) to improve education has broad benefits for scaling personalized learning and teaching, from understanding students' learning status to automating instructional decisions \cite{vincent2020trustworthy, nguyen2023ethical}.
Under this trend, deep learning-based Knowledge Tracing (KT) has become a prominent research area, demonstrating superior performance in modeling students' knowledge mastery \cite{piech2015deep, liu2025deep}. 
KT aims to predict a student's future performance over time from previous learning history. 

However, AI systems may introduce unwanted risk into education \cite{alfredo2024human}, and such concerns are recognized by policymakers. 
The European Union AI Act classifies educational AI models as high-risk, especially those determining access, admission, or evaluating learning outcomes \cite{eu2024aiact}.
The Act requires AI providers to \textit{``perform model evaluation ... with a view to identifying and mitigating systemic risks."}
U.S. Department of Education further emphasizes that AI systems should \textit{``leverage automation to advance learning outcomes while protecting human decision making and judgment”} \cite{office2023artificial}. 
In this regard, Explainable AI(XAI) is key to centering human agency for educational stakeholders and fostering trust by making predictions understandable \cite{khosravi2022explainable}. 

Most XAI works for KT focus on model-based interpretability, either incorporating attention mechanisms or integrating educational psychometric theories \cite{bai2024survey}.
The former inspects the KT model's internal behaviors via attention heatmaps \cite{ghosh2020context, zhao2020interpretable, qin2025interpretable}, while the latter predicts psychometric parameters, such as Item Response Theory \cite{baker2001basics}, and uses those as explanations \cite{chen2023improving, sun2024interpretable, huang2024xkt}. 
While these methods address \textit{`what'} questions, this leaves room to explore \textit{`why?'} and \textit{`how?'} questions \cite{miller2019explanation}. 

In this work, we investigate the potential of counterfactual XAI for KT and propose {\tt KTCF}, a novel counterfactual explanation method for KT. 
Our counterfactual explanations are produced as actionable recourse that is causal, suggesting input changes to achieve the desired outcome \cite{wachter2017counterfactual, ustun2019actionable, karimi2021algorithmic}. 
We believe that counterfactual explanations are suitable for education, as they show higher user satisfaction and trust than other explanation forms \cite{wachter2017counterfactual, warren2024categorical}. 
In KT, an example explanation is \textit{``to change KT model's prediction on knowledge concept (KC) $kc_{10}$ from incorrect to correct, student should change their response on previously incorrect $kc_2$ and $kc_5$."}
This explanation may serve as an educational instruction that guides student actions in learning process.

\begin{figure*}[h!]
\centering
\includegraphics[width=0.99\textwidth]{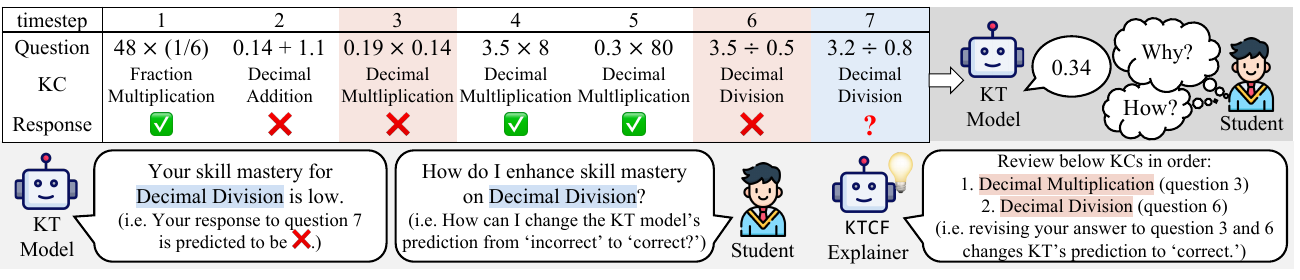}
\caption{An application scenario of our counterfactual explanation for KT under an educational context.}
\label{fig:example}
\end{figure*}

Our contributions of this work are:

\begin{itemize}
    \item We propose a counterfactual explanation generation method for KT that accounts for KC relationships and a post-processing scheme to convert a counterfactual explanation to actionable steps of educational instructions.
    \item We specify conceptualization, problem formulation, and desired properties of generating a counterfactual explanation for KT under the educational context.
\end{itemize}

\section{An Application Scenario}

To illustrate our method’s contribution, we present a hypothetical counterfactual explanation in Figure \ref{fig:example}. 
A student solves a series of problems on \textit{`Decimal Arithmetic,'} and a KT model learns the student's skill mastery based on the student's learning history. 
The KT model predicts the probability that the student will answer correctly on \textit{`Decimal Division'} at timestep $7$ is $0.34$, translated as the student's skill mastery on \textit{`Decimal Division'} is low. 

Rather than asking about what features were relevant or what sections of learning history the KT model focused on for prediction, the student would like to know how to enhance the mastery of \textit{`Decimal Division,'} translated as a counterfactual question to achieve the desired outcome: \textit{``how can I change the KT model's prediction to correct?"}

Our {\tt KTCF} method identifies a sequence of incorrect KCs to be corrected in the student's previous learning history, which translates to indicating changes to the student's response that would flip the KT's prediction to the desired outcome. 
Thus, this explanation can guide the learning process according to the student's needs. 

\subsubsection{In Relation to Educational Theories} 
Our approach is grounded by Bloom's Mastery Learning \cite{block1971mastery}.
The theory operationalizes teaching as initial instruction followed by formative assessment. 
Based on the diagnosis, correction or review procedures are given to failed students so that misunderstandings are not propagated.
Bloom states,
\begin{quote}
\textit{Students respond best when diagnosis is accompanied by specific prescription of alternative instruction materials and processes they can use to overcome their learning difficulties} \cite{bloom1968learning}.
\end{quote}

Bloom demonstrated the effects of Mastery Learning versus conventional classes (teacher-student ratio 1:30), and 1 -1 tutoring. 
Results show that tutored students' achievements are 2 sigma above the conventional class average \cite{bloom19842}. 
Bloom termed the `2 Sigma Problem,' which aims to find ways to replicate the effect of 1-1 tutoring for students. 

KT models may perform as a highly scalable diagnostic tool for measuring skill mastery, and our explanation method could potentially support effective correction or review procedures, resembling some aspects of 1-1 tutoring.
Within this context, it is conceivable that KT and our explanation method could serve as a potential solution to the 2-sigma problem. 

\section{Related Works}

\subsection{Counterfactual Explanation for Education}

While counterfactual explanation is widely studied in domains such as medicine, finance, and process monitoring \cite{wang2021counterfactual, wang2023sparsity, huang2021counterfactual}, its applications on education are relatively underexplored.

Counterfactual explanations for education have been applied to predicting student performance, detecting at-risk students, and analyzing dropout patterns. 
\citeauthor{afrin2023exploring} uses Diverse Counterfactual Explanations ({\tt DiCE}) \cite{mothilal2020explaining} to analyze predictions on whether students will pass or fail at their course \cite{afrin2023exploring}.
\citeauthor{tsiakmaki2021case} generates counterfactuals guided by the nearest class prototype for explaining predictions on whether a student will pass or fail the final exam \cite{tsiakmaki2021case}.
\citeauthor{smith2022individualized} generates counterfactuals by perturbing predictors generated by SHAP for explaining whether a student is at risk of failing a course \cite{smith2022individualized}.
\citeauthor{zhang2023visual} generates and visualizes counterfactuals for analyzing dropout patterns in online learning \cite{zhang2023visual}.

For KT, the notion of counterfactual reasoning has been implemented as a means of aiding model prediction, but has not been discussed under the XAI context \cite{wang2023graphca, zhang2023counterfactual, cui2024interpretable}.

\begin{figure*}[ht!]
\centering
\includegraphics[width=0.99\textwidth]{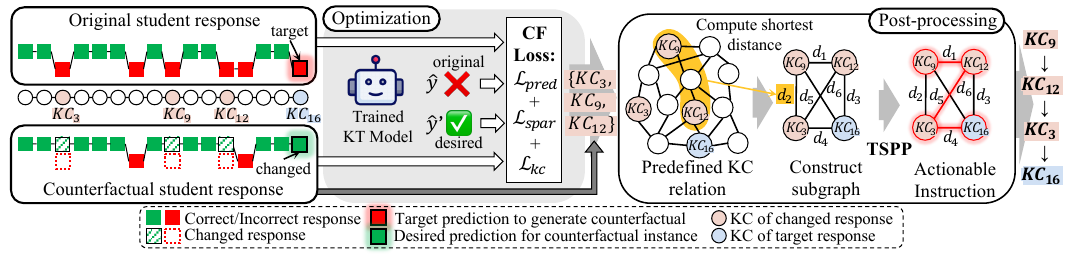}
\caption{An illustration of our proposed {\tt KTCF} method for generating counterfactual explanations for KT.}
\label{fig:algorithm}
\end{figure*}

\subsection{Explainable Knowledge Tracing}

Aforementioned, there exists two threads of XAI research for KT, model-based and post-hoc \cite{bai2024survey}. Model-based explanations for KT are presented as heatmaps of attention weights \cite{ghosh2020context, zhao2020interpretable, qin2025interpretable} and line plots of predicted parameters for psychometric theories \cite{chen2023improving, sun2024interpretable, huang2024xkt}. Post-hoc explanations for KT employ KC maps derived from KT predictions and assess local feature importance using LRP or SHAP \cite{lu2023interpreting, wang2022generic, valero2023shap}.

These forms of explanation allow for inspecting the internal workings of KT models and finding input-output associations. Specifically for local explanations, attention heatmaps can answer \textit{``what part of sequence did the KT model most focus on when making a prediction?"}; psychometric line plots can answer \textit{``does the KT model accurately capture the student’s learning process?"}; and post-hoc feature relevance can answer \textit{``what positive or negative contribution does a KC have on the KT's prediction?"}

While existing methods can provide tools for model behavior inspection and key features analysis, these do not necessarily translate to immediate and actionable educational instruction for a specific, unique student. 
Further, limitations of attention mechanisms for explanations have been addressed in the previous XAI literature \cite{serrano2019attention, bai2021attentions, liu2022rethinking}. 

Notably, \citeauthor{lu2024design} emphasizes stakeholders' agency in using KT for educational support, using Deep SHAP to explain KT prediction as \textit{``$kc_i$ is your weak skill, influenced by your performance on $kc_j, kc_k$”} \cite{lu2024design}. Their user study shows that explaining KT's decisions increases trust, credibility, and knowledge for students and teachers.

\section{Counterfactual Explanations for KT}

\subsection{Problem Conceptualization}

Guided by current advice on XAI research \cite{freiesleben2023dear, langer2021we, miller2019explanation}, we believe that ``\textit{explanations are not just the presentation of associations and causes, they are contextual}" \cite{miller2019explanation}.

We first specify what definition we follow for interpretability and formulate the goal, concepts, and purpose of counterfactual explanations in education.

We follow the definition of \citeauthor{murdoch2019definitions}, 
\begin{quote}
    \textit{We define interpretable machine learning as extraction of relevant knowledge from a machine-learning model concerning relationships either contained in data or learned by the model}
    \cite{murdoch2019definitions}.
\end{quote}

Primary goal is of our explanation is to make KT model predictions understandable to educational stakeholders, which are students, teachers, families and caregivers.

For concepts, we propose \textit{explanandum} (i.e., what is to be explained) and \textit{explanans} (i.e., what we want to explain) for our work \cite{miller2019explanation}. 
We define explanandum for KT as \textit{``how the student learning history would have to be different for the KT model to predict high mastery of a specific KC?"} and explanans as \textit{``students' previous incorrect KCs to be corrected.”}

\subsection{Problem Formulation}

Let $X^{\text{orig}}=[(kc_{1}, r^{\text{orig}}_{1}), (kc_{2}, r^{\text{orig}}_{2}), \cdots,(kc_{t}, r^{\text{orig}}_{t})]$ be a original sequence of a student's learning history up to time $T$, and $R^{\text{orig}}=[r^{\text{orig}}_{1}, r^{\text{orig}}_{2}, \cdots, r^{\text{orig}}_{t}]$ be a sequence of student's responses of $X$. 
Let $f:X^{\text{orig}} \rightarrow [0, 1]$ be a trained KT model that predicts whether a student will give a correct answer to a KC at timestep $t$. 
We focus on a target KC at specific $t$, denoted as $kc_{\text{target}}$, where student's response is incorrect and KT model predicted the student response to be incorrect ($\hat{y}_t = f(X^{\text{orig}}_{t}) = 0$). 
$kc_{\text{target}}$ is the last KC in this sequence. 
Our objective is to find a counterfactual responses $R^{\text{cf}}=[r^{\text{cf}}_{1}, r^{\text{cf}}_{2}, \cdots, r^{\text{cf}}_{t}]$ that constitutes a counterfactual learning history $X^{\text{cf}}=[(kc_{1}, r_{1}^{\text{cf}}), (kc_{2}, r_{2}^{\text{cf}}), \cdots,(kc_{t}, r_{t}^{\text{cf}})]$ such that KT model predicts positive ($\hat{y}_t = f(X^{\text{cf}}) = 1$) for a specific $kc_{\text{target}}$.

\subsection{Properties of Counterfactual Explanations for KT}

We propose properties that a `good' counterfactual explanation should have under educational context:
\begin{itemize}
    \item \textbf{Intervention Sparsity}: Counterfactual explanations should suggest minimum changes as well as be close to the original learning history \cite{wachter2017counterfactual}.
    \item \textbf{Actionability}: Counterfactual explanations should include actionable changes. It is unrealistic to suggest changing KCs to incorrect when the student already answered them correctly \cite{wachter2017counterfactual, verma2024counterfactual}.
    \item \textbf{KC Level Granularity}: Explanations should be framed in terms of KCs, not question items. A good explanation would suggest \textit{``improve your `Decimal Division' skill..."} rather than \textit{``answer `Question 17' correctly..."}
    \item \textbf{KC Relationship Coherence}: Explanations should consider the relationship between KCs. For instance, it would be implausible to suggest practicing \textit{`Integer Addition'} skill to a student solving \textit{`Algebraic Equations.'}
    \item \textbf{Form of Discrete, Sequential Steps of Actions}: Counterfactual explanations should be presented in a series of actions that would guide students from the current decision to the desired decision, complying with the purpose of algorithmic recourse \cite{verma2024counterfactual}.
\end{itemize}

\subsection{Methodology}

Our {\tt KTCF} method serves as a local, post-hoc outcome explanation, formulated as an optimization problem. 
The overview of our method is illustrated in Figure \ref{fig:algorithm}.

\subsubsection{Counterfactual Explanation}

Given a student's original response $R=[r^{\text{orig}}_{1}, r^{\text{orig}}_{2}, \cdots, r^{\text{orig}}_{t}]$ from a learning history $X^{\text{orig}}=[(kc_{1}, r^{\text{orig}}_{1}), (kc_{2}, r^{\text{orig}}_{2}), \cdots,(kc_{t}, r^{\text{orig}}_{t})]$, we initialize our counterfactual response $R^{\text{cf}}$.
Then, we generate our counterfactual explanation through a stochastic optimization using {\tt Adam} \cite{kingma2014adam}.

\begin{algorithm}[tb]
\caption{{\tt KTCF}: Counterfactual Explanation for KT}
\label{alg:generation}
\textbf{Input}: A student's learning history $X^{\text{orig}}$, target KC $kc_{\text{target}}$, trained KT model $f$, KC relation graph $G_{\text{kc}}$ \\
\textbf{Parameter}: $\lambda_{\text{spar}}, \lambda_{\text{kc}}$, max iterations $N_{\text{iter}}$, learning rate $\eta$, early stopping threshold $\tau$ \\
\textbf{Output}: An counterfactual response $R^{\text{cf}}$ derived from the original response $R^{\text{orig}}$
\begin{algorithmic}[1] 
\STATE Define mask $\mathbf{m}$ where $m_t \gets \mathbb{I}_{r_t \in R^{\text{orig}}} [r^{\text{orig}}_t = 0]$ for all $t$
\STATE Initialize $R^{\text{cf}}$ 
\FOR{iteration from 1 to $N_{\text{iter}}$}
    \STATE $\mathcal{L}_{\text{{\tt KTCF}}} \gets \mathcal{L}_{\text{pred}}+\lambda_{\text{spar}} \cdot \mathcal{L}_{\text{spar}} + \lambda_{\text{kc}} \cdot \mathcal{L}_{\text{kc}}$
    \STATE Compute gradient: $\mathbf{\theta} \leftarrow \nabla_{R^{\text{cf}}} \mathcal{L}_{\text{{\tt KTCF}}}$
    \STATE Update counterfactual: $R^{\text{cf}} \gets R^{\text{cf}} - \eta \cdot \mathbf{\theta}$
    \STATE Project onto actionable elements: $R^{\text{cf}} \gets R^{\text{cf}} \odot \mathbf{m} + R^{\text{orig}} \odot (1-\mathbf{m})$
    \IF {$\mathcal{L}_{\text{{\tt KTCF}}}<\tau$}
    \STATE \textbf{break}
    \ENDIF
\ENDFOR
\STATE \textbf{return} $R^{\text{cf}}$
\end{algorithmic}
\end{algorithm}

Our loss function is defined as:
\begin{equation}
    \mathcal{L}_{\text{{\tt KTCF}}}=\mathcal{L}_{\text{pred}} + \lambda_{\text{spar}} \cdot \mathcal{L}_{\text{spar}} + \lambda_{\text{kc}} \cdot \mathcal{L}_{\text{kc}} \text{,}
\end{equation}
where our prediction loss $\mathcal{L}_{\text{pred}}$ is binary cross entropy between predicted probability on a counterfactual instance $X^{\text{cf}}$ and the desired probability of 1.0,
\begin{equation} 
    \begin{split}
    \mathcal{L}_{\text{pred}} = -\log(f(X^{\text{cf}})) \text{,}
    \end{split}
\end{equation}
and sparsity loss $\mathcal{L}_{\text{spar}}$ is Hamming distance between a original response $R^{\text{orig}}$ and the counterfactual response $R^{\text{cf}}$,
\begin{equation} 
    \mathcal{L}_{\text{spar}} = \sum_{t=1}^{T} \mathbb{I}(R^{\text{orig}}_t \neq R^{\text{cf}}_t) \text{.}
\end{equation}

\begin{algorithm}
\caption{Post-processing for Sequential Actions}
\label{alg:post-processing}
\textbf{Input:} Original response sequence $R^{\text{orig}}$, counterfactual response sequence $R^{\text{cf}}$, KC sequence $[kc_1, kc_2,\cdots,kc_t]$, KC relation graph $G_{\text{kc}} = (V_{\text{kc}}, E)$, target KC $kc_{\text{target}}$. \\
\textbf{Output:} Path $H'$ derived from $R^{\text{cf}}$
\begin{algorithmic}[1]
\STATE $\mathcal{I} \gets \{i \mid R^{\text{orig}} \neq R^{\text{cf}}\}$
\STATE $\mathcal{S}_{\text{kc}} \gets \{kc_i \mid i \in \mathcal{I}\}$ 
\STATE $\overline{V}^{\text{CF}} \gets \mathcal{S}_{\text{kc}} \cup \{kc_{\text{target}}\}$ 
\STATE Initialize distance matrix $D \in \mathbb{R}^{|\overline{V}^{\text{CF}} | \times |\overline{V}^{\text{CF}} |}$
\FOR{each $kc_i \in \overline{V}^{\text{CF}}$}
    \FOR{each $kc_j \in \overline{V}^{\text{CF}}$}
        \STATE $D[kc_i, kc_j] \gets$ \textsc{Dijkstra}$(G_{kc}, kc_i, kc_j)$ 
    \ENDFOR
\ENDFOR
\STATE Construct subgraph $G'_{\text{kc}} = (\overline{V}^{\text{CF}}, E', D)$, where $E' = \{(kc_i, kc_j) \mid kc_i, kc_j \in \overline{V}^{\text{CF}}, kc_i \neq kc_j\}$ and edge weights are $D[kc_i, kc_j]$
\STATE $H \gets$ \textsc{Greedy}$(G'_{\text{kc}}, kc_{\text{target}})$
\STATE $H' \gets$\textsc{Rev}$(H)$ 
\RETURN $H'$
\end{algorithmic}
\end{algorithm}

To ensure that counterfactual explanations suggest KC changes that are pedagogically sound and closely related to the $kc_{\text{target}}$, we introduce a penalty term, KC loss $\mathcal{L}_{\text{kc}}$, based on path distance in the loss function. 
We utilize an undirected graph of predefined KC relations, $G_{\text{kc}}=(V_{\text{kc}},E)$, where $V_{\text{kc}}$ is a node set where each node is a KC, and $E$ is an edge set where each edge $e \in E$ indicates a relationship exists between node $KC_{i}$ and node $KC_{j}$. 

The KC loss $\mathcal{L}_{\text{kc}}$ aims to penalize changes that are distant from the $kc_{\text{target}}$ in the KC relation graph, as measured by the shortest path distance $d$. 
Formally, the $\mathcal{L}_{\text{kc}}$ is defined as,
\begin{equation}
    \mathcal{L}_{\text{kc}} = \sum_{kc_i \in \overline{V}^{\text{CF}}} d(kc_i, kc_{\text{target}})
\end{equation}
where $\overline{V}^{\text{CF}}$ denotes KCs modified in the counterfactual.
This guides the optimization to penalize changes to KCs less relevant to the $kc_{\text{target}}$, abiding by the KC relationship coherence.

We apply an actionability mask $\mathbf{m}$ as a crucial step in our process to ensure only the student's originally incorrect answers can be modified.
It generates counterfactuals that are only actionable and enforces complete actionability. 
The full explanation generation process is described in Algorithm \ref{alg:generation}.

\subsubsection{Post-processing for Sequential Actions}

After identifying a set of counterfactual KCs, we convert the explanation to a sequence of educational instructions using a variation of the Traveling Salesman Path Problem (TSPP) \cite{lam2008traveling}. 
Our approach is described in Algorithm \ref{alg:post-processing}.

Given $G_{\text{kc}}$, the goal is to find a Hamiltonian path starting from the $kc_{\text{target}}$ node that traverses all unique nodes in $\overline{V}^{\text{CF}}$.
First, we calculate the shortest path distance between every node pair of our counterfactual KC set $\overline{V}^{\text{CF}}$ using Dijkstra's algorithm.
Second, we create a new, smaller complete graph $G'_{\text{kc}} = (\overline{V}^{\text{CF}}, E', D)$ where edge weights $D$ are the shortest path distance.
Then, we find a Hamiltonian path $H^{*}$ starting from the target KC in a greedy manner on $G'_{\text{kc}}$. The inversed path is provided as a sequence of actionable steps. 

The complexity of Algorithm 2 is $O(|\overline{V^{\text{CF}}}|^2(|V_{\text{kc}}|+|E|)\text{log}(|V_{\text{kc}}|))$. It is small because {\tt KTCF} is optimized for sparsity. 
Under an educational context, this generates a sequential learning instruction that minimizes overall study burden of students.

\begin{table*}[ht!]
\renewcommand{\arraystretch}{1.3}
\fontsize{9pt}{9pt}\selectfont
\centering
\begin{tabularx}{0.95\textwidth}{c|ccccccc}
\hline
\multicolumn{1}{>{\centering\arraybackslash}m{20mm}|}{Methods} & \multicolumn{1}{>{\centering\arraybackslash}m{20mm}}{Validity($\uparrow$)} & \multicolumn{1}{>{\centering\arraybackslash}m{20mm}}{Sparsity($\downarrow$)} & \multicolumn{1}{>{\centering\arraybackslash}m{20mm}}{Sparsity Rate($\downarrow$)} & \multicolumn{1}{>{\centering\arraybackslash}m{20mm}}{Actionability($\downarrow$)} & \multicolumn{1}{>{\centering\arraybackslash}m{20mm}}{Actionability Rate($\downarrow$)} & \multicolumn{1}{>{\centering\arraybackslash}m{20mm}}{Time($\downarrow$)}\\
[1ex]
\hline
{\tt Wachter-rand}  & 0.725$\pm$0.45 & 67.355$\pm$10.33 & 0.338$\pm$0.05 & 40.525$\pm$10.90 & 0.617$\pm$0.20 & 3.791$\pm$0.08 \\
{\tt DiCE-rand} & 0.880$\pm$0.33 & 75.587$\pm$14.73 & 0.380$\pm$0.07 & 35.320$\pm$12.24 & 0.504$\pm$0.25 & \underline{2.565 $\pm$1.87} \\

{\tt KTCF-rn}  & \textbf{0.930 $\pm$ 0.26} & \textbf{49.845 $\pm$ 8.44} & \textbf{0.250 $\pm$ 0.04} & \textbf{0.000 $\pm$ 0.00} & \textbf{0.000 $\pm$ 0.00} & 3.024 $\pm$ 0.91\\
{\tt KTCF-rand} & 0.720 $\pm$ 0.45 &  53.985 $\pm$ 9.18 &  0.271 $\pm$ 0.05 & \textbf{0.000 $\pm$ 0.00} &  \textbf{0.000 $\pm$ 0.00} &  4.700 $\pm$ 0.06 \\
{\tt KTCF-sr}  & \textbf{0.930 $\pm$ 0.26} &  50.075 $\pm$ 7.71 &  0.252 $\pm$ 0.04 & \textbf{0.000 $\pm$ 0.00} & \textbf{0.000 $\pm$ 0.00} &  3.065 $\pm$ 1.15\\
{\tt KTCF-cc} & 0.685 $\pm$ 0.47 & 55.635 $\pm$ 8.31 &  0.280 $\pm$ 0.04 & \textbf{0.000 $\pm$ 0.00} & \textbf{0.000 $\pm$ 0.00} & 3.772 $\pm$ 0.97\\
{\tt KTCF-gs} & \underline{0.920 $\pm$ 0.27} & \underline{49.920 $\pm$ 7.74} & \underline{0.251 $\pm$ 0.04} & \textbf{0.000 $\pm$ 0.00} & \textbf{0.000 $\pm$ 0.00}  &  \textbf{2.202 $\pm$ 1.52}\\
[0.5ex]
\hline
\end{tabularx}
\caption{Evaluation results of counterfacutal explanation generation methods {\tt KTCF}, {\tt Wachter}, and {\tt DiCE} on XES3G5M test data. Best results are shown in bold; second-best are underlined.}
\label{table1}
\end{table*}

\section{Experiment}

We show quantitative and qualitative evaluation of our {\tt KTCF} method.
Since choosing baselines and evaluation criteria for counterfactual explanations heavily depend on types of approaches and the application domain, we select landmark baselines and evaluation metrics from the previous counterfactual XAI literature on benchmarking \cite{mothilal2020explaining, guidotti2024counterfactual, moreira2025benchmarking}.

\subsubsection{Baselines}

We compare our {\tt KTCF} method to two baselines; {\tt Wachter} \cite{wachter2017counterfactual} and {\tt DiCE} \cite{mothilal2020explaining}.
Although being a solid stream of counterfactual XAI, we do not consider Prototype or Nearest Unlike Neighbor approaches \cite{van2021interpretable, delaney2021instance} that we regard the learning process of each student as idiosyncratic \cite{bloom1968learning}.

\subsubsection{Evaluation Metrics}

We choose the following quantitative evaluation metrics: validity, sparsity, sparsity rate, actionability, and generation time. 
To foster evaluation standardization of XAI research, we mention that selected evaluation metrics cover four CO-12 properties: Continuity, Compactness, Context, and Contrastivity \cite{nauta2023anecdotal}.

Given a test dataset of $N$ instances with each instance $X_i$ has length $T$, we define evaluation metrics as follows:

\begin{enumerate}
    \item \textbf{Validity($\uparrow$)} measures the fraction of counterfactuals returned that are actually counterfactuals,
    \begin{equation}
        \text{Validity} = \frac{\sum_{i=1}^{N} \mathbb{I}[f(X^{\text{cf}}_i) > 0.5]}{N}.
    \end{equation}
    \item \textbf{Sparsity($\downarrow$)} measures the number of features that have changed,
    \begin{equation}
        \text{Sparsity} = \frac{1}{N} \sum_{i=1}^{N} \sum_{t=1}^{T} \mathbb{I}({R^{\text{orig}}}^{(i)}_{t} \neq {R^{\text{cf}}}^{(i)}_{t}).
    \end{equation}
    \item \textbf{Sparsity Rate($\downarrow$)} measures the ratio of features that have changed relative to the total number of features,
    \begin{equation}
        \text{Sparsity Rate} = \frac{1}{N} \sum_{i=1}^{N} \frac{\sum_{t=1}^{T} \mathbb{I}({R^{\text{orig}}}^{(i)}_{t} \neq {R^{\text{cf}}}^{(i)}_{t})}{T}.
    \end{equation}
    \item \textbf{Actionability($\downarrow$)} measures unactionable changes in counterfactuals. In KT, an unactionable change occurs when a counterfactual suggests altering a correct response to incorrect,
    \begin{equation}
        \text{Actionability} = \frac{1}{N} \sum_{i=1}^{N} \sum_{t=1}^{T} \mathbb{I}[{R^{\text{orig}}}^{(i)}_{t} = 1 \land {R^{\text{cf}}}^{(i)}_{t} = 0].
    \end{equation}
    \item \textbf{Actionability Rate($\downarrow$)} is the proportion of all changes (Sparsity) that are unactionable,
    \begin{equation}
        \text{Actionability Rate} = \frac{\text{Actionability}}{\text{Sparsity}}.
    \end{equation}
    \item \textbf{Generation Time($\downarrow$)} measures the time (in seconds) the algorithm takes to find a counterfactual.
\end{enumerate}

\begin{figure*}[t]
    \centering
     \begin{subfigure}[b]{0.99\textwidth}
         \centering
         \includegraphics[width=\textwidth]{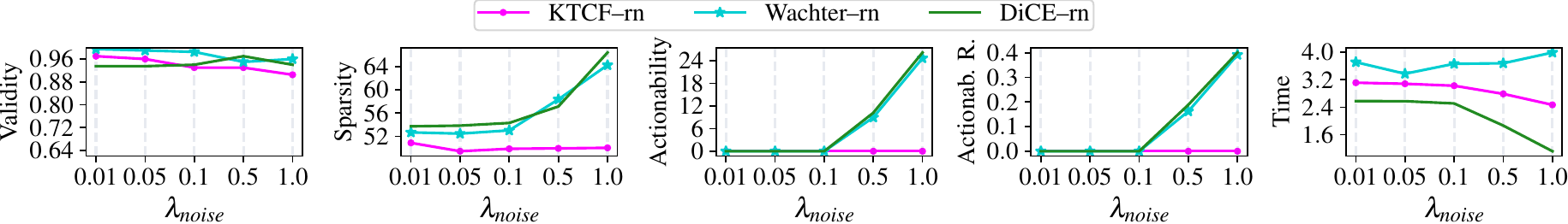}
         \caption{Comparisons of methods with Gaussian noise initialization strategy ({\tt -rn}) with varying hyperparameter $\lambda_{\text{noise}}$.}
         \label{fig:init_noise_std}
     \end{subfigure}
     \begin{subfigure}[b]{0.99\textwidth}
         \centering
         \includegraphics[width=\textwidth]{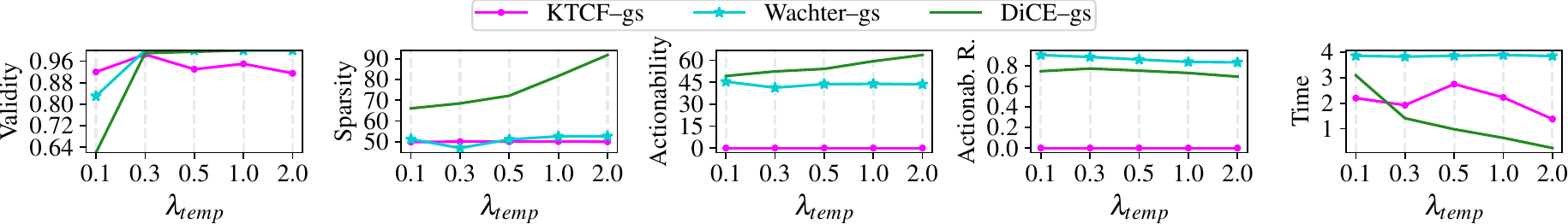}
         \caption{Comparisons of methods with Gumbel-Sigmoid initialization strategy ({\tt -gs}) with varying hyperparameter $\lambda_{\text{temp}}$.}
         \label{fig:init_temp}
     \end{subfigure}
      \begin{subfigure}[b]{0.85\textwidth}
         \centering
         \includegraphics[width=\textwidth]{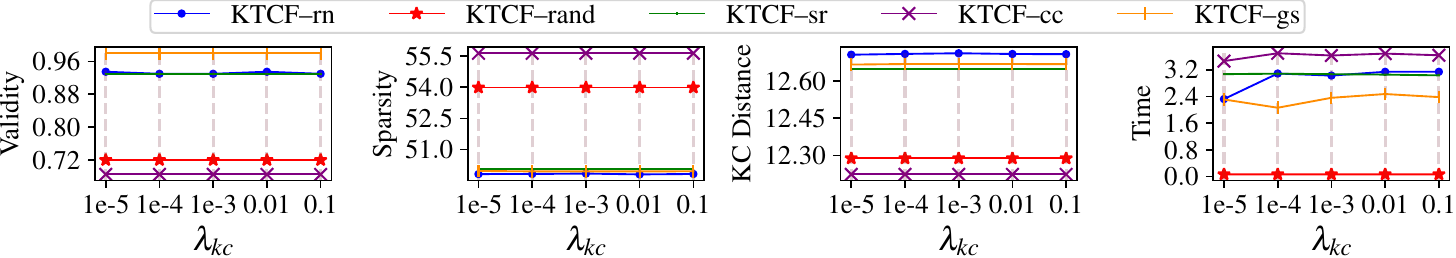}
         \caption{Comparisons of initialization strategies on {\tt KTCF} with varying hyperparameter $\lambda_{\text{kc}}$.}
         \label{fig:lambda_kc}
     \end{subfigure}
    \caption{Ablation study on the sensitivity of {\tt KTCF}, {\tt Wachter}, and {\tt DiCE} methods' performance across hyperparameter variations of initialization strategies {\tt -rn} and {\tt -gs}, and across hyperparameter variations of KC Loss $\lambda_{\text{kc}}$.}
    \label{fig:init_graph}
\end{figure*}

\begin{table*}[t]
\centering
\renewcommand{\arraystretch}{1.2}
\fontsize{9pt}{9pt}\selectfont
\begin{tabular}{>{\centering\arraybackslash}m{15mm}|>{\arraybackslash}m{13cm}|>{\centering\arraybackslash}m{13mm}}
    \hline
    \multicolumn{1}{>{\centering\arraybackslash}m{15mm}|}{Methods} & \multicolumn{1}{>{\centering\arraybackslash}m{13cm}}{Generated Educational Instructions} & \multicolumn{1}{|>{\centering\arraybackslash}m{13mm}}{Total Path Distance}\\
    \hline
    {\tt Wachter} & [Treemap, Ring operation cycle issues, Applications of odd and even numbers, \textcolor{Gray}{Integer difference times}, Find cycle rules in the calendar, \textbf{Distinguish between leap years and non-leap years}, Calendar cycles, Magic square relationship, Simple statistics table, Counting multiples, Inverted type, Prototype questions, Variation questions, Counting units of decimals (deciles, percentiles, tensiles)] & 77\\
    \hline
    {\tt Wachter} \textit{(a)} & \textcolor{Gray}{Integer difference times} $\rightarrow$ Calendar cycles $\rightarrow$ Ring operation cycle issues $\rightarrow$ Magic square relationship $\rightarrow$ Variation questions $\rightarrow$ Prototype questions $\rightarrow$ Inverted type $\rightarrow$ Counting multiples $\rightarrow$ Counting units of decimals (deciles, percentiles, tensiles) $\rightarrow$ Applications of odd and even numbers $\rightarrow$ Simple statistics table $\rightarrow$ Treemap $\rightarrow$ Find cycle rules in the calendar $\rightarrow$ \textbf{Distinguish between leap years and non-leap years} & 66\\
    \hline
    {\tt DiCE} & [Treemap, Treemap, Ring operation cycle issues, Applications of odd and even numbers, Applications of odd and even numbers, Modulo operation, \textcolor{Gray}{How many changes are applied}, \textcolor{Gray}{The problem of two-quantity difference between dark difference type}, Find cycle rules in the calendar, Calendar cycles, Calendar cycles, Even-order magic square filling method, Magic square relationship, Simple statistics table, Counting multiples, Inverted type, Variation questions, Two-volume repulsion, Counting units of decimals (deciles, percentiles, tensiles), \textbf{Distinguish between leap years and non-leap years}] & 95 \\
    \hline
    {\tt DiCE} \textit{(a)} & \textcolor{Gray}{The problem of two-quantity difference between dark difference type} $\rightarrow$ Modulo operation $\rightarrow$ \textcolor{Gray}{How many changes are applied} $\rightarrow$ Calendar cycles $\rightarrow$ Ring operation cycle issues $\rightarrow$ Even-order magic square filling method $\rightarrow$ Magic square relationship $\rightarrow$ Counting units of decimals (deciles, percentiles, tensiles) $\rightarrow$ Applications of odd and even numbers $\rightarrow$ Counting multiples $\rightarrow$ Variation questions $\rightarrow$ Inverted type $\rightarrow$ Two-volume repulsion $\rightarrow$ Simple statistics table $\rightarrow$ Treemap $\rightarrow$ Find cycle rules in the calendar $\rightarrow$ \textbf{Distinguish between leap years and non-leap years} & 79 \\
    \hline
    {\tt KTCF} & [Applications of odd and even numbers, Modulo operation, \textbf{Distinguish between leap years and non-leap years}, Calendar cycles, Counting multiples] & 30\\
    \hline
    {\tt KTCF} \textit{(a)} & Modulo operation $\rightarrow$ Calendar cycles $\rightarrow$ Counting multiples $\rightarrow$ Applications of odd and even numbers $\rightarrow$  \textbf{Distinguish between leap years and non-leap years} & 26 \\
\hline
\end{tabular}
\caption{Comparisons of actionable educational instructions generated by {\tt KTCF}, {\tt Wachter}, and {\tt DiCE} on instance \# 1,452 of the XES3G5M test dataset. Rows annotated with \textit{(a)} indicate results after applying the proposed post-processing scheme. The total path distance is the sum of the shortest paths between each consecutive pair of KCs in the explanation. Target KC $k_{\text{target}}$ is \textbf{bolded} and unactionable changes are indicated in \textcolor{Gray}{gray}. The KC names presented are translated into English for readability; the original texts are in Chinese.}
\label{table2}
\end{table*}

\subsubsection{Counterfactual Initialization}

Generating counterfactuals for categorical features is a known challenge in counterfactual XAI \cite{verma2024counterfactual}.
The challenge is especially severe in KT, where binary student responses make counterfactual quality highly sensitive to initialization.
To empirically investigate this issue, we experiment with five initialization strategies for binary sequences of student responses:

\begin{itemize}
    \item Gaussian noise ({\tt -rn})
        \begin{equation}
          R^{\text{cf}} = R^{\text{orig}} + \lambda_{\text{noise}} \cdot \epsilon, \quad \epsilon_i \sim \mathcal{N}(0, 1)  
        \end{equation}
    \item Random binary ({\tt -rand})
        \begin{equation}
          R^{\text{cf}} = z, \quad z_i \sim \mathrm{Bernoulli}(0.5)  
        \end{equation}
    \item Soft relaxation ({\tt -sr})
        \begin{equation}
          R^{\text{cf}} = \sigma (z), \quad z_i \sim \mathcal{N}(0, 1)  
        \end{equation}
    \item Convex combination ({\tt -cc})
        \begin{equation}
            \begin{aligned}
                R^{\text{cf}} &= \lambda_{\text{cc}} \cdot R^{\text{orig}} + (1 - \lambda_{\text{cc}}) \cdot z, \\  
                z_i \sim &\mathrm{Bernoulli}(0.5), \quad \lambda_{\text{cc}} \in [0, 1]
            \end{aligned}
        \end{equation}
        
    \item Gumbel-Sigmoid Relaxation \cite{jang2016categorical} ({\tt -gs})
        \begin{equation}
            \begin{aligned}
                R^{\text{cf}} &= \sigma\left(\frac{z + g_1 - g_2}{\lambda_{\text{temp}}}\right), \\
                z_i \sim \mathcal{N}(0, 1), \quad& g_1, g_2 \sim \mathrm{Gumbel}(0, 1), \quad \lambda_{\text{temp}} > 0
            \end{aligned}  
        \end{equation}
\end{itemize}

Baselines {\tt Wachter} and {\tt DiCE} follow random initialization({\tt -rand}) as indicated in their works. 

\subsubsection{Dataset} 

We use the XES3G5M \cite{liu2023xes3g5m}, a large-scale educational dataset with auxiliary information on KCs. 
XES3G5M contains 5,549,635 interaction sequences from 18,066 students on Mathematics.
The dataset also provides relationship information among KCs, which we utilize in our {\tt KTCF} method as an undirected graph. 
The constructed KC relation graph has 1,175 nodes and 1,304 edges.
We evaluate the approaches on 200 randomly selected instances from students with more than 45\% incorrect responses.
This selection was made intentionally to focus on students who could benefit from {\tt KTCF} while minimizing bias from students with all-correct responses.

\subsubsection{KT Architecture} 

We use DKT \cite{piech2015deep} as our main KT architecture. 
DKT uses an LSTM network to model student interactions, and shows performance comparable to that of subsequent KT models featuring more advanced and complex architectures \cite{liu2022pykt}. 
For standardizing KT research, we use the {\tt pyKT} library \cite{liu2022pykt} for data preprocessing, training, and evaluation.

\subsubsection{Experimental Setup and KT Performance} 

We run all our experiments on NVIDIA GeForce RTX 4090 devices, and experiment setups are identical to {\tt pyKT}'s setups \cite{liu2022pykt}. 
We perform 5-fold cross-validation on DKT and report the test performance. 
The DKT model demonstrates 0.8415 accuracy and 0.8358 AUC for validation and 0.8253 accuracy and 0.8226 AUC for test data.
For hyperparameters in Algorithm \ref{alg:generation}, we set $\lambda_{\text{spar}}=0.1$, $\lambda_{\text{kc}}=1\mathrm{e}-3$, $N_{\text{iter}}=200$, $\eta$, $\tau=1\mathrm{e}-4$, and $\lambda_{\text{cc}}=0.5$.

\subsection{Experiment Results}

Table \ref{table1} presents our evaluation results. Overall, {\tt KTCF} outperforms baselines across all metrics.
Specifically, {\tt KTCF-rn} improves validity by 28.3\% over {\tt Wachter-rand} and 5.7\% over {\tt DiCE-rand}, and reduces sparsity by 26.0\% and 34.0\%, respectively. 
{\tt KTCF-gs} achieves 41.9\% and 14.2\% faster computation times compared to {\tt Wachter-rand} and {\tt DiCE-rand}. 
Notably, {\tt KTCF} produces only actionable counterfactuals, fully eliminating unactionable modifications. 

Our method generates valid counterfactuals with minimal, actionable changes. 
{\tt KTCF-rn} and {\tt KTCF-gs} are the most balanced in validity, sparsity, actionability, and generation time.
By contrast, {\tt Wachter} and {\tt DiCE} fail to generate valid counterfactuals for 27.5\% and 12\% of students, respectively, and suggest excessive and unactionable changes.

Initialization strategies using Gaussian noise ({\tt -rn}), soft relaxation ({\tt -sr}), and Gumbel-Sigmoid relaxation ({\tt -gs}) yield the best results, suggesting that soft relaxation-based initializations are most effective.
In contrast, hard random ({\tt -rand}) and convex combination ({\tt -cc}) initializations underperform, indicating that hard random starts are ineffective for generating counterfactual explanations for KT.

\subsubsection{Ablation Study}

We investigate how the choice of initialization hyperparameters influences performance, as depicted in Figure \ref{fig:init_noise_std} and \ref{fig:init_temp}. 
We select two initialization strategies {\tt -rn} and {\tt -gs} that demonstrate high performance in our final result, and compare the performance of three methods across hyperparameters $\lambda_{\text{noise}}$ and $\lambda_{\text{temp}}$. 

Overall, our {\tt KTCF} method is highly robust across different hyperparameter values for both initialization strategies. 
{\tt KTCF-rn} and {\tt KTCF-sg} performance in validity, sparsity, and actionability remains consistently high regardless of the noise level and temperature settings.
In contrast, baseline methods are highly sensitive to these hyperparameters. 

To provide intuition for this result, we show the performance of {\tt KTCF} on initialization strategies across ranges of hyperparameter $\lambda_{\text{kc}}$ in Figure \ref{fig:lambda_kc}. 
The results indicate that the KC loss itself strengthens robustness, making performance consistent to initialization strategies across all values of $\lambda_{\text{kc}}$.

\subsubsection{Qualitative Analysis of Actionable Steps}

To evaluate our post-processing scheme for converting counterfactuals to sequential actions, we present an actual instance of generated educational instructions for qualitative comparison in Table \ref{table2}. 
For {\tt KTCF}, we used top performing {\tt KTCF-rn} method to generate instructional steps. 
For {\tt DiCE}, we selected the first explanation among its diverse explanations. 

The total path distance is reduced after applying the post-processing scheme for all methods, indicating that our scheme effectively finds a path that reduces overall study burden of the educational instruction. 
{\tt KTCF} provides the fewest educational instructions with complete actionable suggestions. 
Though average KC distance for {\tt KTCF} are larger ({\tt Wachter}:{\tt DiCE}:{\tt KTCF}=4.71:4.64:5.2), total study burden is lower(66:79:26) due to KTCF’s superior sparsity.

In Table \ref{table2}, KCs presented in {\tt KTCF} are closely related to the target KC of \textit{`Distinguishing between leap and non-leap years.'}
To solve this problem, knowledge on divisibility rule, calendar cycles, modulo operations, and basic integer arithmetic are required, which are included in {\tt KTCF}'s instruction. 
However, instructional steps from {\tt Wachter} and {\tt DiCE} includes seemingly unrelated KCs to \textit{`Distinguishing between leap and non-leap years,'} such as \textit{`Treemap,'} \textit{`Counting units of decimals,'} \textit{`Magic square relationship,'} and \textit{`Simple statics table.'}

\section{Conclusion}

In this work, we propose {\tt KTCF}, a novel method for generating counterfactual explanations in KT as well as method for converting explanation to a sequence of steps for actionable recourse in education. 
Our experiments show that {\tt KTCF} produces high-quality counterfactuals, and our post-processing scheme generates sequential educational instructions to guide learning process.
These findings suggest that our method is effective in providing meaningful explanations that connects KT with its educational purpose and priorities.
Although our method is sensitive to initialization due to the nature of the KT domain, it opens up new possibilities for handling categorical features in counterfactual explanations.
Future work will involve conducting a user study to assess practical impact of {\tt KTCF} on student learning outcomes and the potential of LLMs to convert counterfactual explanations to educational instructions.
Overall, our work contributes to encouraging active discussions on counterfactual XAI for KT and a step towards responsible, stakeholder-centered applications of AI in education.

\section{Acknowledgments}

This research was supported by Basic Science Research Program through the National Research Foundation of Korea(NRF) funded by the Ministry of Education(No.RS-2025-25431740).
This work was supported by the National Research Foundation of Korea(NRF) grant funded by the Korea government(Ministry of Science and ICT)(No.RS-2025-16064585).

\bibliography{aaai2026}

\appendix

\section{Appendix A: Instances of Generated Counterfactual Explanations as Actionable Steps}
\label{apx:instances}

We provide additional examples of generated educational instructions. Table \ref{appendix:table3} presents explanations for test instance \# 2,212, Table \ref{appendix:table4} for test instance \# 1,895, and Table \ref{appendix:table5} for test instance \# 342. In cases where a student solves multiple problems involving the same KC, we specify the problem number alongside the KC to provide unambiguous comparisons across methods. 

\begin{table*}[ht!]
\centering
\renewcommand{\arraystretch}{1.5}
\fontsize{9pt}{9pt}\selectfont
\begin{tabular}{>{\centering\arraybackslash}m{13mm}|>{\arraybackslash}m{13.7cm}|>{\centering\arraybackslash}m{13mm}}
    \hline
    \multicolumn{1}{>{\centering\arraybackslash}m{13mm}|}{Methods} & \multicolumn{1}{>{\centering\arraybackslash}m{13.7cm}}{Generated Educational Instructions} & \multicolumn{1}{|>{\centering\arraybackslash}m{13mm}}{Total Path Distance}\\
    \hline
    {\tt Wachter} & [Addition and disassembly (no requirement), Treemap, Route matching issues, Line and intersection point, Stain counting issues, Arrangement Comprehensive, Geometric Thought Translation, Formula Rectangle Square Perimeter, Formula to find the area of a rectangle, Other types of multiplication principle, Rectangular combo graphic perimeter, Rectangle and square area formula, Rectangle and square area formula, General group number problem, Formula Rectangle Square Perimeter, Other types of multiplication principle, Rectangle and square area formula, Sum of arithmetic sequences, Distribution law such as integer division, Normal type \textit{(question \#  107)}, Comprehensive transformation, \textbf{Formula method to find the area of rectangular square}] & 127 \\
    \hline
    {\tt Wachter} \textit{(a)} & Comprehensive transformation $\rightarrow$ Geometric Thought Translation $\rightarrow$ Rectangle and square area formula $\rightarrow$ Rectangular combo graphic perimeter $\rightarrow$ Formula to find the area of a rectangle $\rightarrow$ Normal type \textit{(question \# 107)} $\rightarrow$ Distribution law such as integer division $\rightarrow$ Sum of arithmetic sequences $\rightarrow$ General group number problem $\rightarrow$ Stain counting issues $\rightarrow$ Other types of multiplication principle $\rightarrow$ Route matching issues $\rightarrow$ Arrangement Comprehensive $\rightarrow$ Line and intersection point $\rightarrow$ Addition and disassembly (no requirement) $\rightarrow$ Treemap $\rightarrow$ Formula Rectangle Square Perimeter $\rightarrow$ \textbf{Formula method to find the area of rectangular square} & 89 \\
    \hline
    {\tt DiCE} & [One stroke application, Addition and disassembly (no requirement), Treemap', Route matching issues, Addition and split number (specified number), Line and intersection point, Stain counting issues, Arrangement Comprehensive, Find the perimeter of trap-type irregular figure, Geometric Thought Translation, Formula Rectangle Square Perimeter, Rectangle and square area formula, Formula to find the area of a rectangle, Geometric method overall reduces blank, Rectangle and square area formula, General group number problem, Formula Rectangle Square Perimeter, Multiplication principle, Other types of multiplication principle, \textcolor{Gray}{Normal type} \textit{(question \# 72)}, Rectangle and square area formula, Formula method to find the area of rectangular square, \textbf{Formula method to find the area of rectangular square}, Sum of arithmetic sequences, Sum of arithmetic sequences, Distribution law such as integer division, Rectangle and square area formula] & 153 \\
    \hline
    {\tt DiCE} \textit{(a)} & Geometric method overall reduces blank $\rightarrow$ Geometric Thought Translation $\rightarrow$ Rectangle and square area formula $\rightarrow$ Find the perimeter of trap-type irregular figure $\rightarrow$ Formula to find the area of a rectangle $\rightarrow$ \textcolor{Gray}{Normal type} \textit{(question \# 72)} $\rightarrow$ Distribution law such as integer division $\rightarrow$ One stroke application $\rightarrow$ Sum of arithmetic sequences $\rightarrow$ General group number problem $\rightarrow$ Stain counting issues $\rightarrow$ Other types of multiplication principle $\rightarrow$ Route matching issues $\rightarrow$ Multiplication principle $\rightarrow$ Arrangement Comprehensive $\rightarrow$ Line and intersection point $\rightarrow$ Addition and split number (specified number) $\rightarrow$ Addition and disassembly (no requirement) $\rightarrow$ Treemap $\rightarrow$ Formula Rectangle Square Perimeter $\rightarrow$ \textbf{Formula method to find the area of rectangular square} & 97 \\
    \hline
    {\tt KTCF} & [Arrangement Comprehensive, \textbf{Formula method to find the area of rectangular square}, Rectangle and square area formula, Geometric method overall reduces blank, Rectangle and square area formula, Formula Rectangle Square Perimeter, Other types of multiplication principle, Rectangle and square area formula, Rectangle and square area formula, Sum of arithmetic sequences, Sum of arithmetic sequences, Normal type \textit{(question \# 107)}, Rectangle and square area formula] & 65 \\
    \hline
    {\tt KTCF} \textit{(a)} & Sum of arithmetic sequences $\rightarrow$ Other types of multiplication principle $\rightarrow$ Arrangement Comprehensive $\rightarrow$ Normal type \textit{(question \# 107)} $\rightarrow$ Geometric method overall reduces blank $\rightarrow$ Rectangle and square area formula $\rightarrow$ Formula Rectangle Square Perimeter $\rightarrow$ \textbf{Formula method to find the area of rectangular square} & 42 \\
\hline
\end{tabular}
\caption{Actionable educational instructions generated on instance \# 2,212 of the XES3G5M test dataset.} 
\label{appendix:table3}
\end{table*}

\begin{table*}[h!]
\centering
\renewcommand{\arraystretch}{1.5}
\fontsize{9pt}{9pt}\selectfont
\begin{tabular}{>{\centering\arraybackslash}m{13mm}|>{\arraybackslash}m{13.7cm}|>{\centering\arraybackslash}m{13mm}}
    \hline
    \multicolumn{1}{>{\centering\arraybackslash}m{13mm}|}{Methods} & \multicolumn{1}{>{\centering\arraybackslash}m{13.7cm}}{Generated Educational Instructions} & \multicolumn{1}{|>{\centering\arraybackslash}m{13mm}}{Total Path Distance}\\
    \hline
    {\tt Wachter} & [Multiple strokes change one stroke, Triangle count, Square, Formula Rectangle Square Perimeter, Find the pass term in arithmetic sequence, Basic problems with solid square matrix, Increase and decrease of solid square matrix, Basic problems with solid square matrix, Basic problems with hollow square matrix, \textcolor{Gray}{Know less and ask for more}, \textcolor{Gray}{Single variable restoration problem}, Practical application, Large amount and multiple, \textbf{Computational solution}] & 80 \\
    \hline
    {\tt Wachter} \textit{(a)} & Large amount and multiple $\rightarrow$ \textcolor{Gray}{Know less and ask for more} $\rightarrow$ Basic problems with hollow square matrix $\rightarrow$ Increase and decrease of solid square matrix $\rightarrow$ Basic problems with solid square matrix $\rightarrow$ \textcolor{Gray}{Single variable restoration problem} $\rightarrow$ Triangle count $\rightarrow$ Find the pass term in arithmetic sequence $\rightarrow$ Multiple strokes change one stroke $\rightarrow$ Formula Rectangle Square Perimeter $\rightarrow$ Square $\rightarrow$ Practical application $\rightarrow$ \textbf{Computational solution} & 62 \\
    \hline
    {\tt DiCE} & [One stroke application, Multiple strokes change one stroke, Ascending (falling), Sum of age problems, Square, Line and intersection point, Formula Rectangle Square Perimeter, Distribution law such as integer division, Find the pass term in arithmetic sequence, Basic problems with solid square matrix, Increase and decrease of solid square matrix, Basic problems with solid square matrix, Increase and decrease of hollow square arrays, Basic problems with hollow square matrix, 2-1-0 Points System, Square matrix problem, \textcolor{Gray}{Single variable restoration problem}, Practical application, Large amount and multiple, \textbf{Computational solution}] & 117 \\
    \hline
    {\tt DiCE} \textit{(a)} & Large amount and multiple $\rightarrow$ Sum of age problems $\rightarrow$ Basic problems with hollow square matrix $\rightarrow$ Increase and decrease of hollow square arrays $\rightarrow$ Increase and decrease of solid square matrix $\rightarrow$ Basic problems with solid square matrix $\rightarrow$ Square matrix problem $\rightarrow$ \textcolor{Gray}{Single variable restoration problem} $\rightarrow$ Ascending (falling) $\rightarrow$ Find the pass term in arithmetic sequence $\rightarrow$ 2-1-0 Points System $\rightarrow$ Line and intersection point $\rightarrow$ Multiple strokes change one stroke $\rightarrow$ One stroke application $\rightarrow$ Distribution law such as integer division $\rightarrow$ Formula Rectangle Square Perimeter $\rightarrow$  Square $\rightarrow$ Practical application $\rightarrow$ \textbf{Computational solution} & 89 \\
    \hline
    {\tt KTCF} & [Multiple strokes change one stroke, Square, Formula Rectangle Square Perimeter, Basic problems with solid square matrix, Increase and decrease of solid square matrix, Basic problems with solid square matrix, Basic problems with solid square matrix, Increase and decrease of hollow square arrays, Practical application, \textbf{Computational solution}] & 37 \\
    \hline
    {\tt KTCF} \textit{(a)} & Increase and decrease of hollow square arrays $\rightarrow$ Increase and decrease of solid square matrix $\rightarrow$ Basic problems with solid square matrix $\rightarrow$ Multiple strokes change one stroke $\rightarrow$ Formula Rectangle Square Perimeter $\rightarrow$ Square $\rightarrow$ Practical application $\rightarrow$ \textbf{Computational solution} & 31 \\
\hline
\end{tabular}
\caption{Actionable educational instructions generated on instance \# 1,895 of the XES3G5M test dataset.} 
\label{appendix:table4}
\end{table*}

\begin{table*}[ht!]
\centering
\renewcommand{\arraystretch}{1.5}
\fontsize{9pt}{9pt}\selectfont
\begin{tabular}{>{\centering\arraybackslash}m{13mm}|>{\arraybackslash}m{13.7cm}|>{\centering\arraybackslash}m{13mm}}
    \hline
    \multicolumn{1}{>{\centering\arraybackslash}m{13mm}|}{Methods} & \multicolumn{1}{>{\centering\arraybackslash}m{13.7cm}}{Generated Educational Instructions} & \multicolumn{1}{|>{\centering\arraybackslash}m{13mm}}{Total Path Distance}\\
    \hline
    {\tt Wachter} & [Treemap, The periodic problem of basic arrangement, The periodic problem of basic arrangement, Cycle issues, Integer number of good friends, Distribution law such as integer division, Two quantity and times, Sum of age problems, Single object bar chart, Odd-order magic square filling method, Variation questions, Understanding of decimals, \textcolor{Gray}{Sum of arithmetic sequences}, \textcolor{Gray}{Find the pass term in arithmetic sequence} \textit{(question \# 180)}, \textbf{Find the number of terms in arithmetic sequence}] & 77 \\
    \hline
    {\tt Wachter} \textit{(a)} & Two quantity and times $\rightarrow$ Sum of age problems $\rightarrow$ Odd-order magic square filling method $\rightarrow$ The periodic problem of basic arrangement $\rightarrow$ Cycle issues $\rightarrow$ Integer number of good friends $\rightarrow$ Understanding of decimals $\rightarrow$ Distribution law such as integer division $\rightarrow$ Treemap $\rightarrow$ Single object bar chart $\rightarrow$ Variation questions $\rightarrow$ \textcolor{Gray}{Find the pass term in arithmetic sequence} \textit{(question \# 180)} $\rightarrow$ \textcolor{Gray}{Sum of arithmetic sequences} $\rightarrow$ \textbf{Find the number of terms in arithmetic sequence} & 74 \\
    \hline
    {\tt DiCE} & [Treemap, The rules of addition and subtraction between odd and even numbers, The periodic problem of basic arrangement, Applications of odd and even numbers, Two quantity and times, Single object bar chart, Variation questions, Understanding of decimals, Letters represent numbers, Find the pass term in arithmetic sequence \textit{(question \# 158)}, \textbf{Find the number of terms in arithmetic sequence}] & 65 \\
    \hline
    {\tt DiCE} \textit{(a)} & Two quantity and times $\rightarrow$ The periodic problem of basic arrangement $\rightarrow$ Letters represent numbers $\rightarrow$ Understanding of decimals $\rightarrow$ Applications of odd and even numbers $\rightarrow$ The rules of addition and subtraction between odd and even numbers $\rightarrow$ Treemap $\rightarrow$ Single object bar chart $\rightarrow$ Variation questions $\rightarrow$ Find the pass term in arithmetic sequence \textit{(question \# 158)} $\rightarrow$ \textbf{Find the number of terms in arithmetic sequence} & 57  \\
    \hline
    {\tt KTCF} & [The periodic problem of basic arrangement, Cycle issues, Two quantity and times, Variation questions, \textbf{Find the number of terms in arithmetic sequence}] & 23 \\
    \hline
    {\tt KTCF} \textit{(a)} & Two quantity and times $\rightarrow$ The periodic problem of basic arrangement $\rightarrow$ Cycle issues $\rightarrow$ Variation questions $\rightarrow$ \textbf{Find the number of terms in arithmetic sequence} & 21 \\
\hline
\end{tabular}
\caption{Actionable educational instructions generated on instance \# 342 of the XES3G5M test dataset.} 
\label{appendix:table5}
\end{table*}

\newpage

\section{Appendix B: On Complete Test Data}

In our experiment, we used a test set of students with incorrect responses of more than 45\%. 
Our choice is intentional to only regard students who might benefit from KTCF and to avoid bias from students with all-correct responses. 
To demonstrate that using the complete test data may introduce bias, we reproduced Table 1 with the complete test data as shown in Table \ref{table1_completetest}. Overall performance improves, likely reflecting students who excel. 

\begin{table*}[ht!]
\renewcommand{\arraystretch}{1.3}
\fontsize{9pt}{9pt}\selectfont
\centering
\begin{tabularx}{0.95\textwidth}{c|ccccccc}
\hline
\multicolumn{1}{>{\centering\arraybackslash}m{20mm}|}{Methods} & \multicolumn{1}{>{\centering\arraybackslash}m{20mm}}{Validity($\uparrow$)} & \multicolumn{1}{>{\centering\arraybackslash}m{20mm}}{Sparsity($\downarrow$)} & \multicolumn{1}{>{\centering\arraybackslash}m{20mm}}{Sparsity Rate($\downarrow$)} & \multicolumn{1}{>{\centering\arraybackslash}m{20mm}}{Actionability($\downarrow$)} & \multicolumn{1}{>{\centering\arraybackslash}m{20mm}}{Actionability Rate($\downarrow$)} & \multicolumn{1}{>{\centering\arraybackslash}m{20mm}}{Time($\downarrow$)}\\
[1ex]
\hline
{\tt Wachter-rand} & 0.646$\pm$0.48 & 43.473$\pm$15.40 & 0.218$\pm$0.08 & 20.762$\pm$8.84 & 0.494$\pm$0.18 & 2.919$\pm$0.08 &  \\
{\tt DiCE-rand} & 0.897$\pm$0.30 & 41.593$\pm$18.61 & 0.209$\pm$0.09 & 11.386$\pm$7.61 & 0.283$\pm$0.20 & \underline{1.488$\pm$1.15} &  \\
{\tt KTCF-rn} & \textbf{0.970$\pm$0.17} & \textbf{20.738$\pm$12.43} & \textbf{0.104$\pm$0.06} & \textbf{0.000$\pm$0.00} & \textbf{0.000$\pm$0.00} & 1.729$\pm$0.72 &  \\
{\tt KTCF-rand} & 0.667$\pm$0.47 & 22.652$\pm$13.45 & 0.114$\pm$0.07 & \textbf{0.000$\pm$0.00} & \textbf{0.000$\pm$0.00} & \textbf{0.041$\pm$0.01} &  \\
{\tt KTCF-sr} & \underline{0.957$\pm$0.20} & \underline{20.949$\pm$12.46} & \underline{0.105$\pm$0.06} & \textbf{0.000$\pm$0.00} & \textbf{0.000$\pm$0.00} & 1.711$\pm$0.85 &  \\
{\tt KTCF-cc} & 0.670$\pm$0.47 & 23.563$\pm$13.79 & 0.118$\pm$0.07 & \textbf{0.000$\pm$0.00} & \textbf{0.000$\pm$0.00} & 1.942$\pm$0.84 &  \\
{\tt KTCF-gs} & 0.910$\pm$0.29 & 21.000$\pm$12.45 & 0.106$\pm$0.06 & \textbf{0.000$\pm$0.00} & \textbf{0.000$\pm$0.00} & 1.583$\pm$0.85 &  \\
[0.5ex]
\hline
\end{tabularx}
\caption{Evaluation results of counterfacutal generation methods {\tt KTCF}, {\tt Wachter}, and {\tt DiCE} on the \textit{complete} XES3G5M test dataset with DKT model.}
\label{table1_completetest}
\end{table*}

\section{Appendix C: Additional KT Models}

We further experimented with our {\tt KTCF} on DKVMN and SAKT models. For test data, DKVMN showed 0.8226 accuracy and 0.8173 AUC. The SAKT demonstrated 0.8158 accuracy and 0.8002 AUC. The experiment results of {\tt KTCF} on these two models are presented in Table \ref{table1_dkvmn} and Table \ref{table1_sakt}. For both models, {\tt KTCF} achieves the best and the second-best performance on major metrics, consistent with experiment results on DKT in Table \ref{table1}. 

\begin{table*}[ht!]
\renewcommand{\arraystretch}{1.3}
\fontsize{9pt}{9pt}\selectfont
\centering
\begin{tabularx}{0.95\textwidth}{c|ccccccc}
\hline
\multicolumn{1}{>{\centering\arraybackslash}m{20mm}|}{Methods} & \multicolumn{1}{>{\centering\arraybackslash}m{20mm}}{Validity($\uparrow$)} & \multicolumn{1}{>{\centering\arraybackslash}m{20mm}}{Sparsity($\downarrow$)} & \multicolumn{1}{>{\centering\arraybackslash}m{20mm}}{Sparsity Rate($\downarrow$)} & \multicolumn{1}{>{\centering\arraybackslash}m{20mm}}{Actionability($\downarrow$)} & \multicolumn{1}{>{\centering\arraybackslash}m{20mm}}{Actionability Rate($\downarrow$)} & \multicolumn{1}{>{\centering\arraybackslash}m{20mm}}{Time($\downarrow$)}\\
[1ex]
\hline
{\tt Wachter-rand}   & 0.920$\pm$0.27 & 65.785$\pm$8.08 & 0.331$\pm$0.04 & 11.695$\pm$5.19 & 0.178$\pm$0.07 & 12.050$\pm$0.87 &  \\
{\tt DiCE-rand}   & \underline{0.985$\pm$0.12} & 68.537$\pm$16.60 & 0.344$\pm$0.08 & 4.003$\pm$4.31 & 0.050$\pm$0.06 & \underline{1.089$\pm$2.29} &  \\
{\tt KTCF-rn}   & 0.930$\pm$0.26 & \underline{48.670$\pm$7.69} & \underline{0.245$\pm$0.04} & \textbf{0.000$\pm$0.00} & \textbf{0.000$\pm$0.00} & 8.900$\pm$2.56 &  \\
{\tt KTCF-rand}   & 0.845$\pm$0.36 & 53.990$\pm$8.78 & 0.271$\pm$0.04 & \textbf{0.000$\pm$0.00} & \textbf{0.000$\pm$0.00} & \textbf{0.187$\pm$0.04} &  \\
{\tt KTCF-sr}   & 0.940$\pm$0.24 & 49.900$\pm$7.73 & 0.251$\pm$0.04 & \textbf{0.000$\pm$0.00} & \textbf{0.000$\pm$0.00} & 10.164$\pm$3.52 &  \\
{\tt KTCF-cc}   & 0.930$\pm$0.26 & 55.635$\pm$8.31 & 0.280$\pm$0.04 & \textbf{0.000$\pm$0.00} & \textbf{0.000$\pm$0.00} & 8.617$\pm$2.79 &  \\
{\tt KTCF-gs}   & \textbf{0.995$\pm$0.07} & \textbf{45.525$\pm$7.74} & \textbf{0.229$\pm$0.04} & \textbf{0.000$\pm$0.00} & \textbf{0.000$\pm$0.00} & 4.389$\pm$2.47 &  \\
[0.5ex]
\hline
\end{tabularx}
\caption{Evaluation results of counterfacutal generation methods {\tt KTCF}, {\tt Wachter}, and {\tt DiCE} on XES3G5M test dataset with DKVMN model.}
\label{table1_dkvmn}
\end{table*}

\begin{table*}[ht!]
\renewcommand{\arraystretch}{1.3}
\fontsize{9pt}{9pt}\selectfont
\centering
\begin{tabularx}{0.95\textwidth}{c|ccccccc}
\hline
\multicolumn{1}{>{\centering\arraybackslash}m{20mm}|}{Methods} & \multicolumn{1}{>{\centering\arraybackslash}m{20mm}}{Validity($\uparrow$)} & \multicolumn{1}{>{\centering\arraybackslash}m{20mm}}{Sparsity($\downarrow$)} & \multicolumn{1}{>{\centering\arraybackslash}m{20mm}}{Sparsity Rate($\downarrow$)} & \multicolumn{1}{>{\centering\arraybackslash}m{20mm}}{Actionability($\downarrow$)} & \multicolumn{1}{>{\centering\arraybackslash}m{20mm}}{Actionability Rate($\downarrow$)} & \multicolumn{1}{>{\centering\arraybackslash}m{20mm}}{Time($\downarrow$)}\\
[1ex]
\hline
{\tt Wachter-rand}    & 0.635$\pm$0.48 & 68.175$\pm$8.83 & 0.343$\pm$0.04 & 13.925$\pm$5.72 & 0.204$\pm$0.08 & \underline{13.092$\pm$19.65} &  \\
{\tt DiCE-rand}   & \underline{0.940$\pm$0.24} & 75.285$\pm$16.50 & 0.378$\pm$0.08 & 8.202$\pm$6.48 & 0.099$\pm$0.08 & \textbf{4.632$\pm$10.67} &  \\
{\tt KTCF-rn}   & 0.825$\pm$0.38 & \textbf{48.730$\pm$8.04} & \textbf{0.245$\pm$0.04} & \textbf{0.000$\pm$0.00} & \textbf{0.000$\pm$0.00} & 33.120$\pm$11.03 &  \\
{\tt KTCF-rand}   & 0.750$\pm$0.43 & 54.110$\pm$8.35 & 0.272$\pm$0.04 & \textbf{0.000$\pm$0.00} & \textbf{0.000$\pm$0.00} & 31.826$\pm$12.61 &  \\
{\tt KTCF-sr}   & 0.895$\pm$0.31 & 50.305$\pm$7.52 & 0.253$\pm$0.04 & \textbf{0.000$\pm$0.00} & \textbf{0.000$\pm$0.00} & 34.132$\pm$34.05 &  \\
{\tt KTCF-cc}   & 0.605$\pm$0.49 & 55.645$\pm$8.30 & 0.280$\pm$0.04 & \textbf{0.000$\pm$0.00} & \textbf{0.000$\pm$0.00} & 33.140$\pm$11.36 &  \\
{\tt KTCF-gs}   & \textbf{0.980$\pm$0.14} & \underline{50.240$\pm$7.74} & \underline{0.252$\pm$0.04} & \textbf{0.000$\pm$0.00} & \textbf{0.000$\pm$0.00} & 32.751$\pm$9.74 &  \\
[0.5ex]
\hline
\end{tabularx}
\caption{Evaluation results of counterfacutal generation methods {\tt KTCF}, {\tt Wachter}, and {\tt DiCE} on XES3G5M test dataset with SAKT model.}
\label{table1_sakt}
\end{table*}

\end{document}